\documentclass{article}

\usepackage[english]{babel}
\usepackage{multirow}
\usepackage{adjustbox}
\usepackage{lineno}
\usepackage[letterpaper,top=2cm,bottom=2cm,left=3cm,right=3cm,marginparwidth=1.75cm]{geometry}

\usepackage{amsmath}
\usepackage{amsfonts}
\usepackage{bm}
\usepackage{graphicx}
\usepackage[colorlinks=true, allcolors=blue]{hyperref}

\title{Fredholm Integral Equations Neural Operator (FIE-NO) \\ for Data-Driven Boundary Value Problems}
\author{Haoyang Jiang, Yongzhi Qu\\
Mechanical Engineering, The University of Utah, 1495 E 100 S, Salt Lake City, 84112}
\date{}

\begin{document}
\maketitle
\begin{abstract}
In this paper, we present a novel Fredholm Integral Equation Neural Operator (FIE-NO) method, an integration of Random Fourier Features and Fredholm Integral Equations (FIE) into the deep learning framework, tailored for solving data-driven Boundary Value Problems (BVPs) with irregular boundaries. Unlike traditional computational approaches that struggle with the computational intensity and complexity of such problems, our method offers a robust, efficient, and accurate solution mechanism, using a physics inspired design of the learning structure. We demonstrate that the proposed physics-guided operator learning method (FIE-NO) achieves superior performance in addressing BVPs. Notably, our approach can generalize across multiple scenarios, including those with unknown equation forms and intricate boundary shapes, after being trained only on one boundary condition. Experimental validation demonstrates that the FIE-NO method performs well in simulated examples, including Darcy flow equation and typical partial differential equations such as the Laplace and Helmholtz equations. The proposed method exhibits robust performance across different boundary conditions. Experimental results indicate that FIE-NO achieves higher accuracy and stability compared to other methods when addressing complex boundary value problems with varying numbers of interior points.
\end{abstract}
\section{Introduction}

In the landscape of computational physics and engineering, machine learning algorithms like Physics-Informed Neural Networks (PINNs) \cite{pinn} and neural operators \cite{NO,boulleoperator} have marked a significant paradigm shift in solving partial differential equations (PDEs) \cite{bvplearning,gkn}. PINNs integrate physical laws into the learning process, effectively constraining neural networks to respect to the underlying physics of the problem. Neural operators, for example, DeepONet \cite{deeponet}, extended the capabilities of neural networks beyond regression between variables and demonstrated its ability to learn mappings between function spaces represented by vectored inputs and outputs. Neural operator has been shown the capability to to address complex problem, such as parameterized PDEs \cite{NO}. Moreover, recent studies have further expanded the scope of neural operators. For instance, the Fourier Neural Operator (FNO) \cite{FNO} has focused on optimizing neural operator architectures to learn a general solution for a class of PDEs with the same structure. \cite{salvi2021neural} enhances the capabilities of the Neural Operator Network by combining recurrent neural network into operator learning, tailoring it for stochastic PDEs and integrating noise with dynamic driving forces. 

In a classic Boundary Value Problem (BVP), the information given includes a known differential equation defining the system's characteristics and the conditions that the solution must satisfy on the boundaries \cite{bvp_book}. In a data-driven setting, the underlying differential equations may be unknown. The underlying governing equations are specified by measurement data. Boundary Integration Equations (BIEs) reformulate these conditions into integral equations focused on the boundary, streamlining the solution process, especially for complex and irregular boundaries. Therefore, this work is inspired by the idea of BIEs can be incorporated into the process of solving BVPs.


However, the incorporation of BIEs presents several significant challenges. First of all, traditional methods are computational expensive due to the need for recalculations when boundary conditions change and the inherent complexity of solving these equations for intricate boundary geometries \cite{beno}.  A more efficient and adaptive approach, capable of handling complex BVP, is desired. Second, the integration of machine learning, such as neural operators and PINNs, with BIEs, has been a significant challenge towards overcoming these hurdles, since the complexity of these models increases significantly with the intricacy of the boundary conditions and the geometry involved. Simplifying these processes is important for practical and scalable applications in computational fields.


To address the above challenges, this paper introduces a Fredholm Integral
Equations (FIEs) operator learning (FIE-NO) method for tackling BVP. The proposed method can solve a data-driven BVP problem with unknown equation forms and irregular boundary shapes. In particular, it's designed to be able to generalize to multiple scenarios after being trained on a single boundary condition type, offering a versatile and robust solution to complex boundary value problems. This approach integrates the mathematical structure into deep learning, addressing the challenges of irregular boundaries and equation uncertainty in a unified framework.


\section{Preliminaries}

\subsection{Boundary Value Problems}

Boundary Value Problems aims to find a solution to a differential equation subject to certain boundary conditions. BVP can be reformulated with BIE over the boundaries to simplify the computational process.  However, the forms of BIE can be complex and change with the specific partial differential equation and the geometry of the domain \cite{BIEbook}. 

In the study of Boundary Value Problems (BVPs), understanding the types of boundary conditions is essential. Two primary types of boundary conditions are:

\textbf{Dirichlet Boundary Conditions:} These conditions specify the values that the solution must take on the boundary of the domain. Formally, if $\Omega$ denotes the domain and $\partial\Omega$ its boundary, then a Dirichlet condition can be expressed as $u(x) = g(x)$ for $x \in \partial\Omega$, where $u(x)$ is the solutions on the boundary, $g(x)$ is a given function.
    
\textbf{Neumann Boundary Conditions:} These conditions specify the values of the derivative of the solution normal to the boundary. They can be formally written as $\frac{\partial u}{\partial n}(x) = h(x)$ for $x \in \partial\Omega$, where $\frac{\partial u}{\partial n}$ denotes the derivative of $u$ in the direction normal to the boundary $\partial\Omega$, and $h(x)$ is a specified function.

These conditions are crucial in defining the solution's behavior at the domain's boundaries and significantly influence the analytical or numerical approach to solving BVPs.

\subsection{Random Fourier Features}

Kernel methods are crucial in machine learning algorithms, allowing operations in a high-dimensional feature space without explicitly computing coordinates. Computing the kernel for every pair of points in large datasets is computationally expensive. Random Fourier Features (RFF) provide an efficient approximation of kernel functions, particularly in large-scale applications.

RFF maps input data into a lower-dimensional Euclidean space where the inner product approximates the kernel function. Based on Bochner’s theorem, for \( x, y \in \mathbb{R}^d \), a translation-invariant and positive definite kernel function \( k(x, y) \) can be represented as the inverse Fourier transform of its Fourier transform:

\begin{equation}
k(x - y) = \int_{\mathbb{R}^d} p(\omega) e^{j \omega^\top (x - y)} d\omega
\end{equation}

Here, \( p(\omega) \) is non-negative and is the Fourier transform of \( k \). It can be interpreted as a probability density function from which samples can be drawn. RFF approximates \( k(x, y) \) using a finite number of samples from \( p(\omega) \), allowing effective estimation of kernel functions without directly computing the entire kernel matrix.

\subsection{Fredholm Integral Equations}

Fredholm Integral Equations (FIE) is an integral equation widely applied in mathematics, physics, and engineering, especially for problems involving partial differential equations (PDEs)\cite{FIE}. BIE can be converted into an FIE and vice-versa \cite{BIEbook}. 

There are two primary forms of Fredholm Integral Equations:

\textbf{First Kind:}

\begin{equation}
f(x) = \int_{S} K(x, t) \varphi(t) \, dS(t)
\end{equation}

where, \( t\) is a dummy variable for integration, and \( K(x, t) \) is the kernel function. Here, the kernel function \( K (s, t)\) and \( f (x)\) are given to find the unknown function \( \varphi (x)\), and the equation involves an integral over the entire boundary \( S \) with the kernel \( K(x, t) \).

\textbf{Second Kind:}

\begin{equation}
\varphi(x) = g(x) + \lambda \int_{S} K(x, t) \varphi(t) \, dS(t)
\end{equation}

where, \( \lambda \) is a constant,  and \( S \) represents the boundary over which the integration is performed. This form includes an additional term \( g(x) \), which can represent boundary conditions or other known functions. In a typically setting, the function \( g (x)\) and \( K (x, t)\) are given to find the unknown function  \( \varphi(x) \).

\section{Methodology}

In this work, we focus on solving data-driven Boundary Value Problems (BVPs) characterized by Dirichlet boundary conditions, which specify the values of the solution on the boundary of the domain. Many physical and engineering problems, such as heat conduction in a solid object and fluid flow in a porous medium, fall into this category. We will also show an extension to Neumann boundary conditions, which specify the derivative of the solution normal to the boundary in section 3.3.

\subsection{Problem Formulation}

Given an unknown underlying function \( f \) defined over a domain \( \Omega \) with boundary \( \partial\Omega \), let \( B \subset \partial\Omega \) represent the known boundary and \( I \subset \Omega \) represent the interior points. We are provided with boundary data \( (t_b, \varphi(t_b)) \) and limited interior data \( (x_i, \varphi(x_i)) \). The subscript \( b \) denotes complete boundary information. From this boundary information, we randomly sample a fixed number of points as input. Our objective is, given boundary data \( (t_b, \varphi(t_b)) \) and \( x_i \), to learn a mapping \( F : (x_i, t_b, \varphi(t_b)) \mapsto \varphi(x_i) \). The learned underlying function \( F \) enables us to solve BVP problems under arbitrary boundary conditions, specifically focusing on those of the Dirichlet where the solution values are directly specified on the boundary.

To address this problem, we propose a Fredholm Integral Equations Neural Operator (FIE-NO) method. The proposed method models the boundary value problems using Fredholm Integral Equations (FIE) of the second kind in section 2.3, which can handle Dirichlet boundary conditions effectively. The FIE-NO method is designed to learn the mapping from boundary conditions to the PDE solutions and use Random Fourier Features (RFF) to approximate the kernel functions.

We formulate the BVP problem as an optimization problem to learn an operator that maps boundary conditions to PDE solutions. This optimization includes minimizing a cost function that reflects the accuracy of the operator in predicting solutions of boundary value problems, thus solving PDEs within irregular domains. The optimization problem is given as:

\begin{equation}
\min_{F} \sum_{(x_i, \varphi(x_i))} \| F(t_b) - \varphi(x_i) \|^2
\end{equation}

subject to any physical or geometric constraints of the problem.

This approach leverages the power of machine learning to predict the behavior of complex systems without requiring an explicit form of the equations, thereby generalizing well across different scenarios and potentially reducing the computational burden associated with traditional numerical methods.

\subsection{Fredholm Integral Equations Neural Operator}
\subsubsection{Simplifying Fredholm Integral Equations}

To reduce computational complexity in large-scale problems, we propose to utilize the Random Fourier Features (RFF) method described in Section 2.3 to approximate the kernel function. Since the kernel function \( K(x, t) \) is unknown, we assume it to be a Gaussian kernel, a reasonable assumption due to its stationarity and positive definiteness properties. Assuming \( p(\omega) \) in Eq. (1) follows a Gaussian distribution \( \omega \sim \mathcal{N}(0, \sigma^2 I) \), we sample \( D \) frequency vectors \( \{ \omega_1, \omega_2, \ldots, \omega_D \} \) from this distribution and generate \( D \) bias terms \( \{ b_1, b_2, \ldots, b_D \} \) uniformly from the interval \([0, 2\pi]\).

Using these sampled frequency vectors and bias terms, we construct a feature map \( z(x) \):

\begin{equation}
z(x) = \sqrt{\frac{2}{D}} \left[ \cos(\omega_1^\top x + b_1), \cos(\omega_2^\top x + b_2), \ldots,
\cos(\omega_D^\top x + b_D) \right]
\end{equation}

Thus, the kernel function \( K(x, t) \) can be approximated as the inner product of the feature maps:

\begin{equation}
K(x, t) \approx z(x)^\top z(t)
\end{equation}

Using RFF to approximate the kernel function simplifies the computational process in the Fredholm integral equation:

\begin{equation}
\varphi(x) = g(x) + \frac{\lambda}{D} \sum_{i=1}^D \cos(\omega_i^\top x + b_i) \int_{S} \cos(\omega_i^\top t + b_i) \varphi(t)dS(t)
\end{equation}

This formulation retains the essence of the Fredholm Integral Equation while leveraging the computational efficiency of RFF. To further simplify the estimation of Eq. 3, we assume $\varphi(t)$, the function inside the integral of can be expressed as a sum of trigonometric functions. A detailed derivation is given in Appendix A. An overall flow is given in Eq. 9 - Eq. 13.

First, let $\varphi(t)$ be represented as:

\begin{equation}
    \varphi(t) = \sum_{n} a_n \cos(nt) + b_n \sin(nt),
\end{equation}

where $a_n$ and $b_n$ are the coefficients corresponding to the cosine and sine components, respectively. Substituting this expression into the integral term of Eq. 7, we get:

\begin{equation}
\begin{aligned}
    \int_{S} \cos(w_i^\top t + b_i) \varphi(t) dS(t) &= \int_{S} \bigg(\cos(w_i^\top t + b_i)\bigg) \times \left( \sum_{n} a_n \cos(nt) + b_n \sin(nt) \right) dS(t).
\end{aligned}
\end{equation}

Second, by decomposing \( \cos(w_i t + b_i) \) within the integral as:

\begin{equation}
\cos(w_i t + b_i) = \cos(w_i t)\cos(b_i) - \sin(w_i t)\sin(b_i)
\end{equation}

The integral then becomes:

\begin{equation}
\begin{aligned}
    \int_{S} \Big( &\cos(w_i t)\cos(b_i) - \sin(w_i t)\sin(b_i) \Big) \cdot \left( \sum_n a_n \cos(n t) + b_n \sin(n t) \right) dS(t)
\end{aligned}
\end{equation}

For simplicity, we define $ c = \frac{1}{2} (a_{w_i} \cos(b_i) - b_{w_i} \sin(b_i)) $, and exploiting the orthogonality of trigonometric functions, the final integral expression simplifies to:

\begin{equation}
    \int_{S} \cos(w_i^\top t + b_i) \varphi(t) dS(t) = c \int_{S} dS(t),
\end{equation}

Consequently, the Fredholm integral equation with the application of RFF becomes:

\begin{equation}
    \varphi(x) = g(x) + \frac{\lambda}{D} \sum_{i=1}^{D} \cos(w_i^\top x) c \int_{S} dS(t).
\end{equation}

This formulation retains the essence of the FIE while leveraging the computational efficiency of RFF, particularly applicable for data-driven BVP.

\subsubsection{Neural Network Approximation}

\begin{figure}[h]
\centering
\includegraphics[width=1\linewidth]{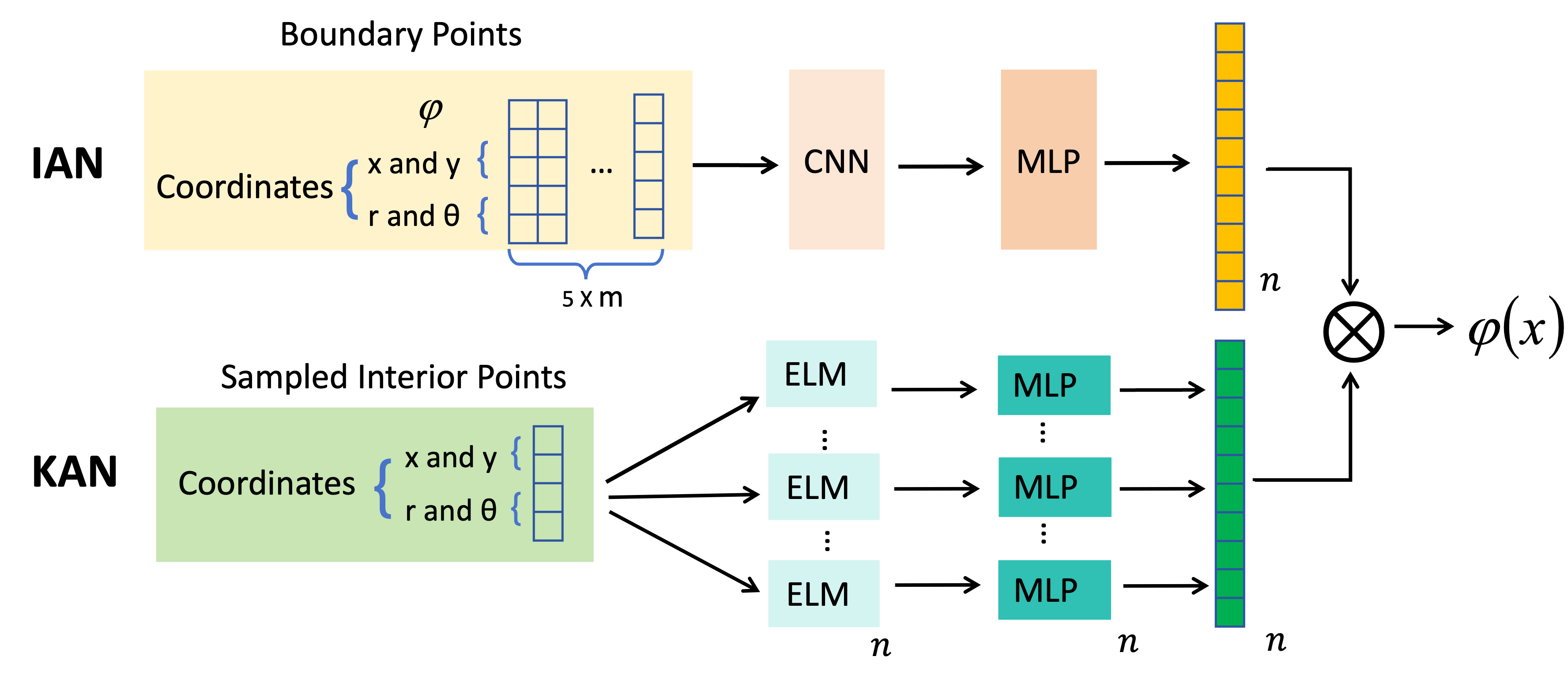}
\caption{\label{fig:model}Model Structure: The diagram shows the overall model architecture, including the processing of boundary and interior points. Boundary points are processed through CNN to extract features, which are then transformed by MLP. Interior points are processed through ELM-inspired layers, followed by MLP with cosine activation and GELU activation. IAN and KAN handle integral and kernel approximations, with their outputs multiplied and summed to produce values for the interior points. For each training session, \(m\) = 200 boundary points are sampled along the boundary, including boundary values $\varphi(x)$, Cartesian coordinates $(x, y)$, and polar coordinates $(r, \theta)$. These inputs are processed together through a Convolutional Neural Network (CNN) to extract features, which are then transformed by a MLP. Interior points include only their Cartesian and polar coordinates $(x, y)$ and $(r, \theta)$. These coordinates are processed through multiple Extreme Learning Machine (ELM)-inspired hidden layers with fixed weights, followed by a MLP network with cosine activation function and a learnable GELU activation function. The Integral Approximation Network (IAN) and Kernel Approximation Network (KAN) handle the integral and kernel approximations. Their outputs are multiplied and summed to produce the values for the interior points. }
\end{figure}

Considering the simplified form of the FIE in the context of neural operator learning, we will design a new neural network framework to approximate the simplified FIE given in Eq. 13. In this study, we proposed a neural network based model comprising two independent components: the Kernel Approximation Network (KAN) and the Integral Approximation Network (IAN). An overall model structure is shown in Fig.\ref{fig:model}.

\textbf{Extreme Learning Machine (ELM) for Kernel Approximation Network (KAN)}

The KAN is tasked with learning a set of basis functions that approximate the kernel of our target integral equation. It autonomously identifies the most fitting representations of the kernel from the training data. Each basis function corresponds to particular features in the input space, and KAN aims to discover the optimal basis functions.

The $KAN(w_i^\top, x)$ in the FIE can be expressed as the cosine terms with randomly sampled parameters:

\begin{equation}
    KAN(w_i^\top, x) \approx  \cos(w_i^\top x),
\end{equation}

where $w_i$ are vectors randomly sampled from a predefined distribution. These parameters are sampled from a standard normal distribution with mean 0 and standard deviation 1, ensuring the stationarity and positive definiteness properties of the Gaussian kernel. This random feature approximation lends itself naturally to the framework of an Extreme Learning Machine (ELM). ELM, known for its ability to handle random features effectively, can be employed to approximate these basis functions terms. In an ELM setup, the input layer is transformed by these random features. The outputs from ELM will be processed by a multiple layer perceptron (MLP) and the MLP weights  are the only trainable parameters. This structure is particularly advantageous for our problem because it directly aligns with the random nature of $w_i$ in the kernel approximation. 

\textbf{Neural Network for Integral Approximation Network (IAN)}

Concurrently, IAN operates in parallel, focusing on learning the integral computation. Rather than processing outputs from KAN, IAN is dedicated to approximating the integral component of the equation based on the boundary function $\varphi(t)$ and the domain $S$. A Convolutional Neural Network (CNN) and a MLP are used for the approximation. The integral part $c \int_{S} dS(t)$ is handled by a separate standard neural network, this network is designed to approximate the integral values.



\subsubsection{Combined Model and Implementation}

 IAN operates in parallel with the KAN, each network learning to approximate different aspects of the integral equation. Eventually, their learned features and weights are combined multiplicatively to approximate the solution of the entire integral equation. This operation mirrors the relationship between the kernel and the integral computation in operator theory. This method enhances the model's capability to understand and address the equation's complexity, improving the precision of the approximate solution.

Thus, the solution modelled by our framework is expressed as follows:

\begin{equation}
\varphi_{approx}(x) = g(x) + \lambda \sum_{i=1}^{D} KAN(w_i; x) \cdot IAN(\Phi(t); t)
\end{equation}

where \( \lambda \) is a scaling factor, and in the classic formulation of FIE for BVPs, the function $g(x)$ represents the boundary condition's influence on the solution, where $g$ could, in principle, be a function of both $\varphi(x)$ and $x$, i.e., $g(\varphi(x), x)$. This relation is crucial as it aligns with the nature of BVPs where boundary conditions, represented by $\varphi(x)$, play a pivotal role in shaping the solution across the domain. Drawing parallels with approaches seen in Neural Integral Equations\cite{2022NIE}, $g(x)$ could be envisioned as a neural network that models the dependency of the boundary conditions on $\varphi(x)$ and $x$. 

However, for the sake of simplification and to underscore the effectiveness of our method, we consider $g(x)$ to be zero. This decision is not merely for computational convenience but also to highlight the robustness of our proposed method in tackling BVPs without the explicit need to model the boundary conditions' direct impact. Essentially, we are demonstrating that even in the absence of a direct modeling of $g(x)$ as a function of $\varphi(x)$ and $x$, our approach remains effective, leveraging the Fredholm Integral Equations Neural Operator (FIE-NO) to solve BVPs with Dirichlet boundary conditions effectively.

With the above simplification, the Fredholm integral equation in our context simplifies to:

\begin{equation}
\varphi_{approx}(x) = \lambda \sum_{i=1}^{D} KAN(w_i; x) \cdot IAN(\Phi(t); t)
\end{equation}

This neural network-based integral equation is at the heart of our method, enabling the neural operator to learn the mapping from boundary conditions to the solution within the domain, effectively addressing the complexities associated with Dirichlet boundary conditions in BVPs. 

The incorporation of the ELM for approximating the $\cos(w_i^\top x)$ terms and a standard neural network for the integral part results in a hybrid model. This model captures the complexities of the input transformation through the ELM and the nuances of the integral operator via the neural network. In implementing this model, the ELM should be designed with a sufficiently large number of hidden nodes to capture the complex patterns in the data. The standard neural network can be a multi-layer perceptron, tailored to the specific requirements of the integral approximation task.

This methodology leverages the strengths of both ELM and standard neural networks, offering a new approach to tackle FIE. It holds potential for applications in fields where traditional numerical methods are limited, such as high-dimensional problems and scenarios with complex boundary conditions.





\subsection{Extension to Neumann Boundary Conditions}
 Our initial derivation shown above has been focus on leveraging the Fredholm Integral Equations (FIE) framework to address challenges of Boundary Value Problems (BVPs) associated with Dirichlet boundary conditions. However, the versatility of the proposed approach allows for an effective extension to encompass Neumann boundary conditions as well. Neumann boundary conditions, which specify the derivative of the solution normal to the boundary, are observed in various physical and engineering contexts, representing another critical aspect of many BVPs.

Adapting our methodology to include Neumann conditions entails a modification of the FIE formulation. This adjustment allows for the integration of derivative terms within the boundary conditions, aligning with the specifications of Neumann problems. The modified FIE, representative of Neumann boundary conditions, is formulated as follows:

\begin{equation}
    \varphi(x) = \lambda \int_{\Omega} K(x, y) h(y) dy, \quad x \in \partial\Omega
\end{equation}

In this equation, we have $h(y) = \frac{\partial \varphi(y)}{\partial n}$, where $\lambda$ signifies a scalar constant, and \(K(x, y)\) is the kernel function that delineates the interactions between points \(x\) and \(y\) within the domain \(\Omega\). The term \(\frac{\partial \varphi(y)}{\partial n}\) denotes the derivative of the function \(\varphi(y)\) normal to the boundary, thereby encapsulating the Neumann conditions. Furthermore, \(h(x)\) represents the function that specifies the derivative values along the boundary \(\partial\Omega\), thus completing the representation of second-kind boundary conditions.

With the simplification, the above Fredholm integral equation in Eq. 17 simplifies to:

\begin{equation}
\varphi_{approx}(x) = \lambda \sum_{i=1}^{D} KAN(w_i; x) \cdot IAN(\frac{\partial \varphi(y)}{\partial n}; t)
\end{equation}

This integral equation enables the neural operator to address the complexities associated with Neumann boundary conditions in BVPs. The expansion to Neumann boundary conditions demonstrates the flexibility of the proposed FIE-NO method and significantly enhances its application range. By detailing the approach's adaptability to handle both first-kind (Dirichlet) and second-kind (Neumann) boundary conditions, we underline its comprehensive utility in addressing a wide array of complex BVPs encountered in scientific and engineering domains. The inclusion of this modification solidifies the FIE-NO method as a robust and versatile tool in the computational toolkit for solving BVPs, underscoring its potential to contribute meaningfully across diverse applications.

Furthermore, the subsequent sections will provide detailed computational tests for both first-kind (Dirichlet) and second-kind (Neumann) boundary conditions. These tests aim to demonstrate the efficacy and versatility of our proposed FIE-NO approach in handling diverse types of boundary value problems. Through these examples, we will showcase the model's robustness and accuracy in solving BVPs under varying conditions, thereby emphasizing its practical applicability in complex physical and engineering scenarios.

\section{Test Example}

For comparative analysis, our method was tested against other approaches such as Nonlinear Integral Equation (NIE) and Advanced Nonlinear Integral Equation (ANIE) approaches \cite{2022NIE}, Green Operator (GO) Method \cite{aldirany2024operator} and Green Learning (GL) Method \cite{greenlearning}. The tests and validation involve 3 kinds of PDE (Laplace equation, Helmholtz equation, and Darcy flow equation) and varying sample sizes. We assessed mean square error, the results show that FIE-NO provided better performance in addressing complex BVPs, particularly due to its ability to handle different types of boundary conditions and equation forms effectively.

We also tested other algorithms. For example, the algorithm in \cite{bvplearning} is not suitable for processing dense sampling at the boundary because the algorithm needs to construct a finite element triangle. The triangle constructed when the boundary points are dense will have a very small corner, so the algorithm is limited to the boundary constructed by straight lines, the effect is poor in our example, and the error is more than ten percent, so it is not used as a comparison. For the work \cite{lu2024fast}, it directly use the boundary information as the input of Branchnet, which has a similar neural network construction to us,  however, our construction is from the Boundary integral equations, not directly from the DeepONet and its universal approximation. The results from those comparisons are not included in this paper since they do not generate full comparable results.

\subsection{Example}

We use the 2D Laplace equation, the 2D Helmholtz equation, and the Darcy flow equation as examples. The Laplace equation is as follows:

\begin{align}
\left\{
\begin{array}{ll}
\Delta u(x) = 0 & \text{in } \Omega \\
u(x) = u_0(x) \quad \text{or} \quad \frac{\partial u(x)}{\partial n} = u_0(x) & \text{on } \Gamma
\end{array}
\right.
\end{align}

And the Helmholtz equation with a wave number \( k \) is given by:

\begin{align}
\left\{
\begin{array}{ll}
\Delta u(x) + k^2 u(x) = 0 & \text{ in } \Omega \\
u(x) = u_0(x) \quad \text{or} \quad \frac{\partial u(x)}{\partial n} = u_0(x) & \text{on } \Gamma
\end{array}
\right.
\end{align}

In this text, we set \( k = 1 \).

The Darcy flow equation is as follows:
 
\begin{align}
\left\{
\begin{array}{ll}
\-\nabla \cdot (a(x) \nabla u(x)) = f(x) & \text{ in } \Omega \\
u(x) = u_0(x) \quad \text{or} \quad \frac{\partial u(x)}{\partial n} = u_0(x) & \text{on } \Gamma
\end{array}
\right.
\end{align}

For the training data, we assume the shape of the boundary is the same for all equations, as shown in Fig. \ref{fig:boundary}, which is created using the following equation:

\begin{equation}
\begin{aligned}
    r(\alpha; t) = 1 + 0.2(&t_1\sin(3\alpha) + t_2 \sin(4\alpha) + t_3\sin(6\alpha) + \\
    &t_4\cos(2\alpha) + t_5\cos(5\alpha)), \quad \alpha \in [0, 2\pi).
\end{aligned}
\end{equation}

We can adjust the shape of the boundary by the parameter $t_2 = 1$ and other $t_i = 0$, we can plot the boundary as follows:

\begin{figure}
\centering
\includegraphics[width=0.65\linewidth]{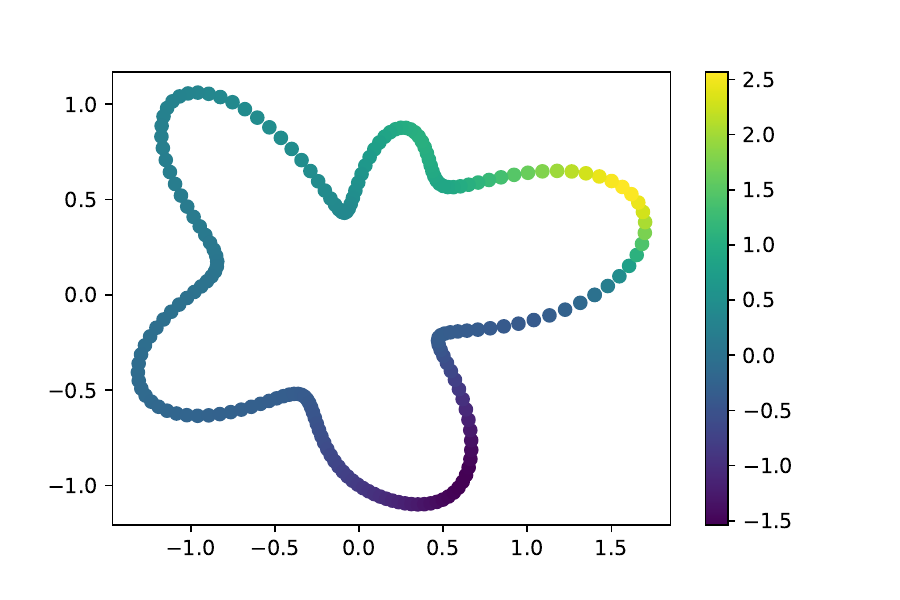}
\caption{\label{fig:boundary}Boundary}
\end{figure}


    

\subsection{Data Preparation}

In the absence of an explicit functional form of the equation and with only the boundary conditions specified, there exists a multitude of potential solutions. Therefore, for effective training of the model, it is necessary to incorporate some interior points in addition to the boundary conditions. This will enable the learning algorithm to accurately infer the equation's form.

Using the predefined boundary in Eq. 22, 200 points are arbitrarily selected with XY coordinates, polar coordinates, and boundary values as the input of IAN in each training batch.  Then we arbitrarily select a total of 50, 100 or 200 points from the interior, containing both XY coordinates and polar coordinates but no values as the input of KAN.

During training, the complete information of the boundary points and the position information of the interior points are used to learn the values of those  interior points through the constructed neural network. The evaluated loss is mean square error (MSE) between the true value and the predicted value. Then, in the testing phase, we verify whether it can generalize to the interior area within different boundaries. It is worth noting that training is done only on one boundary condition, it is then tested for generalization to other boundaries.

\subsection{Neural Network Design}

The IAN architecture is a convolutional neural network (CNN) comprising two 2D convolutional layers with GELU activations and two fully connected layers. Convolution operations can be seen as performing a small interval integration. The convolution process in CNN can approximate the integral calculations over a small interval, which is the reason for using CNN to learn the integral in the proposed methods. The CNN maps to a 20-dimensional output vector after flattening and dimension reduction through the IAN.


KAN includes multiple Extreme Learning Machines (ELM), where the weights are initialized using a standard normal distribution and remain fixed during training, reflecting the principle of random hidden nodes in ELM. A cosine activation function is added after the ELM layers. The results are then transformed through a parameter-learnable non-linear mapping and activated by the GELU function. The output from each MLP are aggregated to form the final output vector of KAN.


\subsection{Results and Discuss}

\begin{figure}[ht]
\centering
\includegraphics[width=1.02\textwidth]{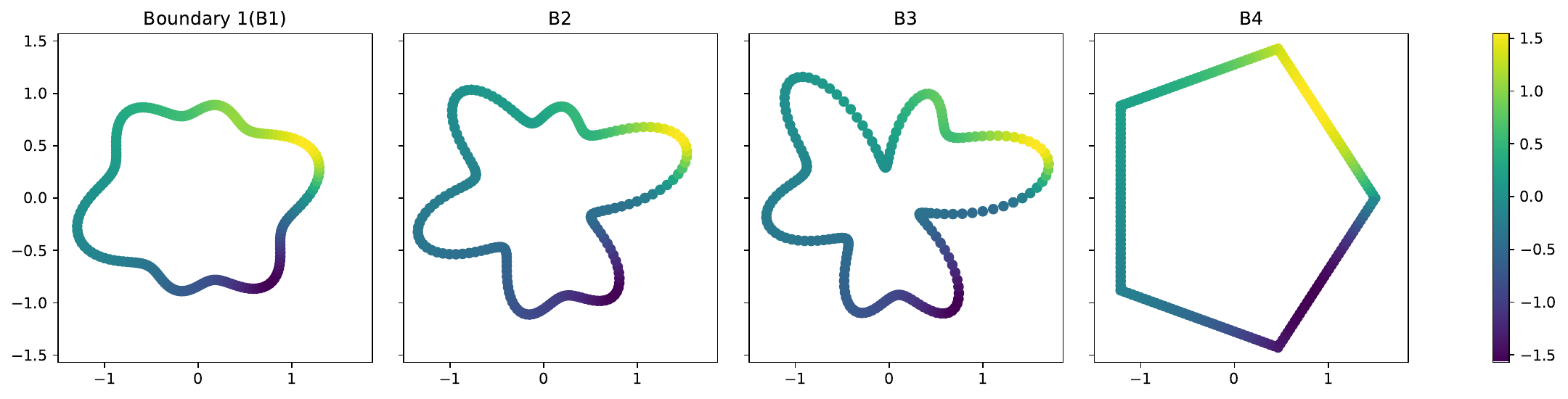} 
\caption{Dirichlet Boundary conditions}
\label{fig:bc1}
\end{figure}

We modified the value of $t_i$ in Eq. 21 arbitrarily to produce three irregular boundaries. In addition to that, we also created a pentagonal boundary. Fig. \ref{fig:bc1} shows the sampling of these Dirichlet boundaries. We subsequently conducted experiments using various equations on these four boundaries.

The shapes of the four boundaries for the  Neumann boundary are set as the same as those in the above set, although the values have been altered for the Neumann boundary. The diagram of these Neumann boundaries is shown below in Fig. \ref{fig:bc2}. Further experiments using different equations were carried out on these four boundaries.

\begin{figure}[ht]
\centering
\includegraphics[width=1.02\textwidth]{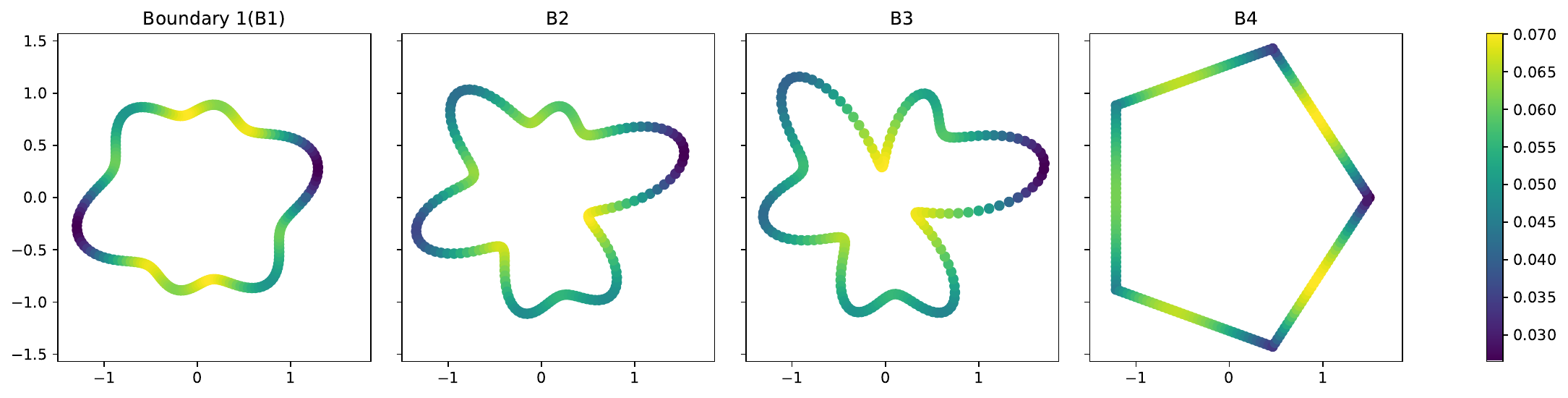} 
\caption{Neumann Boundary conditions}
\label{fig:bc2}
\end{figure}

\subsubsection{2D Laplace equation}

\textbf{Dirichlet Boundary} In this scenario, we set the value of the boundary as 

\begin{equation}
  u_0(x,y) = e^{x} \sin(y)
\end{equation}

For four different boundaries shown in Fig.\ref{fig:bc1}, we get the comparison result as the following in Table. \ref{tab:tab1}

\begin{table}[h]
\centering
\caption{Results with Dirichlet Boundary of Laplace equation(values in units of \(10^{-3}\))}
\begin{adjustbox}{width=\textwidth,center}

\begin{tabular}{|c|c|c|c|c|c|c|c|c|c|c|c|c|c|}
\hline
\multirow{2}{*}{} & \multicolumn{3}{c|}{B1} & \multicolumn{3}{c|}{B2} & \multicolumn{3}{c|}{B3} & \multicolumn{3}{c|}{B4} \\ \cline{2-13}
                  & 50 & 100 & 200 & 50 & 100 & 200 & 50 & 100 & 200 & 50 & 100 & 200 \\ \hline
FIE-NO & \textbf{\textit{0.175}} & \textbf{\textit{0.140}} & \textbf{\textit{0.109}} & 0.120 & 0.078 & 0.047 & \textbf{\textit{0.135}} & 0.094 & \textbf{\textit{0.063}} & \textbf{\textit{0.232}} & \textbf{\textit{0.166}} & \textbf{\textit{0.139}} \\ \hline
NIE & 0.188 & 0.146 & 0.120 & \textbf{\textit{0.115}} & \textbf{\textit{0.073}} & \textbf{\textit{0.041}} & 0.138 & \textbf{\textit{0.093}} & \textbf{\textit{0.063}} & 0.282 & 0.218 & 0.188 \\ \hline
ANIE & 0.192 & 0.151 & 0.121 & 0.129 & 0.081 & 0.048 & 0.149 & 0.104 & 0.079 & 0.234 & 0.168 & 0.140 \\ \hline
GO & 0.198 & 0.177 & 0.144 & 0.137 & 0.096 & 0.065 & 0.156 & 0.107 & 0.083 & 0.295 & 0.224 & 0.191 \\ \hline
GL & 0.201 & 0.180 & 0.149 & 0.143 & 0.109 & 0.081 & 0.184 & 0.133 & 0.102 & 0.308 & 0.239 & 0.198 \\ \hline
\end{tabular}
\end{adjustbox}
\label{tab:tab1}
\end{table}

\begin{figure}[ht]
\centering
\includegraphics[width=1.02\textwidth]{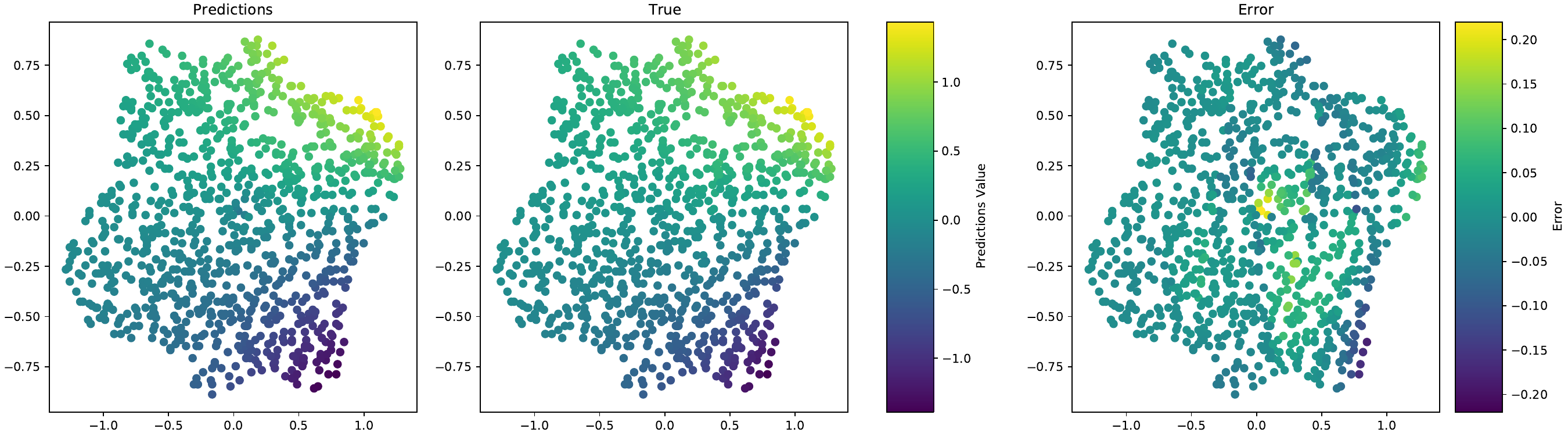} 
\caption{Results}
\label{fig:plot1}
\end{figure}

\begin{figure}[ht]
\centering
\includegraphics[width=1.02\textwidth]{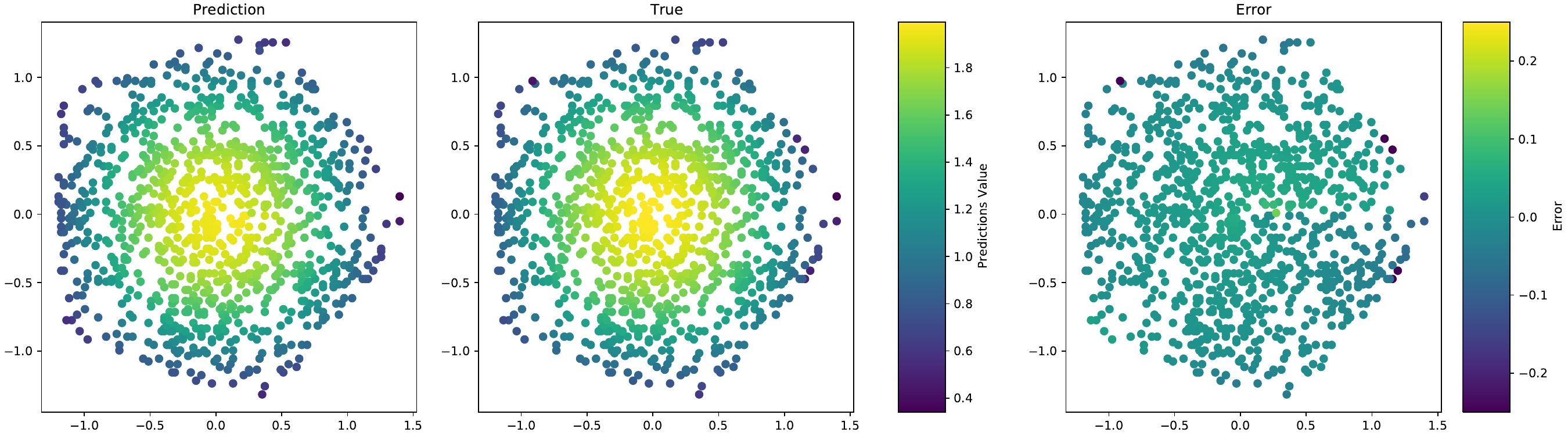} 
\caption{Results}
\label{fig:pplot1}
\end{figure}

We take the second and fourth boundary conditions (B2 and B4) as examples to show the result of our method, the predicted interior points, the true points, and the mean square error are shown in Fig. 
 \ref{fig:plot1} and Fig. \ref{fig:pplot1}.

Table. \ref{tab:tab1} provides a comparative analysis with other methods across varying sample sizes (50, 100, and 200). For B1, B2, and B3, we can find the FIE-NO and the NIE get the best result, their difference is not obvious, but for B4, we can find that FIE-NO is better than other methods, although we assume the boundary is smooth for FIE-NO, and the ANIE's performance surpass the NIE, close to the FIE-NO.  



\textbf{Neumann Boundary} In this scenario, we set the value of the boundary as 

\begin{equation}
\frac{\partial u_0(x,y)}{\partial n} = 0.1 \cos(x) \cos(y)
\end{equation}

For these four different boundaries in Fig.\ref{fig:bc1}, we get the comparison result as the following Table. \ref{tab:tab1}

\begin{table}[h]
\centering
\caption{Results with Neumann Boundary of Laplace equation(values in units of \(10^{-3}\))}
\begin{adjustbox}{width=\textwidth,center}

\begin{tabular}{|c|c|c|c|c|c|c|c|c|c|c|c|c|c|}
\hline
\multirow{2}{*}{} & \multicolumn{3}{c|}{B1} & \multicolumn{3}{c|}{B2} & \multicolumn{3}{c|}{B3} & \multicolumn{3}{c|}{B4} \\ \cline{2-13}
                  & 50 & 100 & 200 & 50 & 100 & 200 & 50 & 100 & 200 & 50 & 100 & 200 \\ \hline
FIE-NO & \textbf{\textit{0.085}} & \textbf{\textit{0.057}} & \textbf{\textit{0.036}} & \textbf{\textit{0.080}} & \textbf{\textit{0.055}} & \textbf{\textit{0.033}} & \textbf{\textit{0.095}} & \textbf{\textit{0.064}} & \textbf{\textit{0.043}} & \textbf{\textit{0.246}} & \textbf{\textit{0.191}} & 
\textbf{\textit{0.162}}\\ \hline
NIE & 0.092 & 0.066 & 0.044 & 0.085 & 0.063 & 0.041 & 0.101 & 0.079 & 0.051 & 0.282 & 0.220 & 0.178 \\ \hline
ANIE & 0.088 & 0.061 & 0.039 & 0.082 & 0.056 & 0.034 & 0.099 & 0.071 & 0.049 & 0.256 & 0.199 & 0.165 \\ \hline
GO & 0.098 & 0.077 & 0.056 & 0.097 & 0.076 & 0.056 & 0.119 & 0.090 & 0.067 & 0.275 & 0.224 & 0.188 \\ \hline
GL & 0.099 & 0.080 & 0.059 & 0.103 & 0.082 & 0.061 & 0.114 & 0.083 & 0.062 & 0.280 & 0.229 & 0.197 \\ \hline
\end{tabular}
\end{adjustbox}
\label{tab:tab2}
\end{table}

\begin{figure}[ht]
\centering
\includegraphics[width=1.02\textwidth]{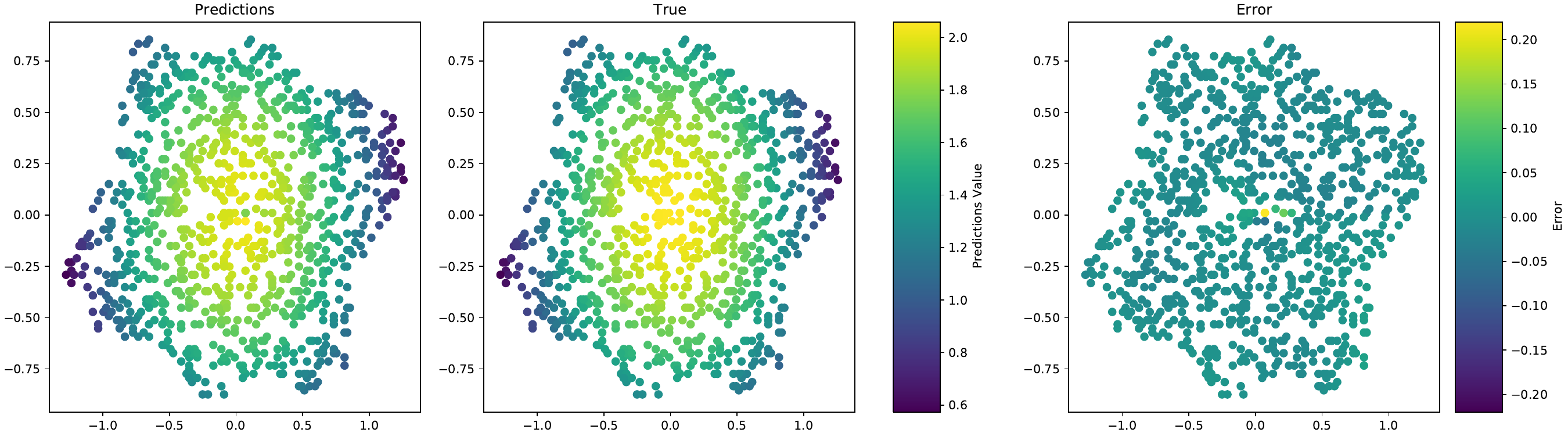} 
\caption{Results}
\label{fig:plot2}
\end{figure}

\begin{figure}[ht]
\centering
\includegraphics[width=1.02\textwidth]{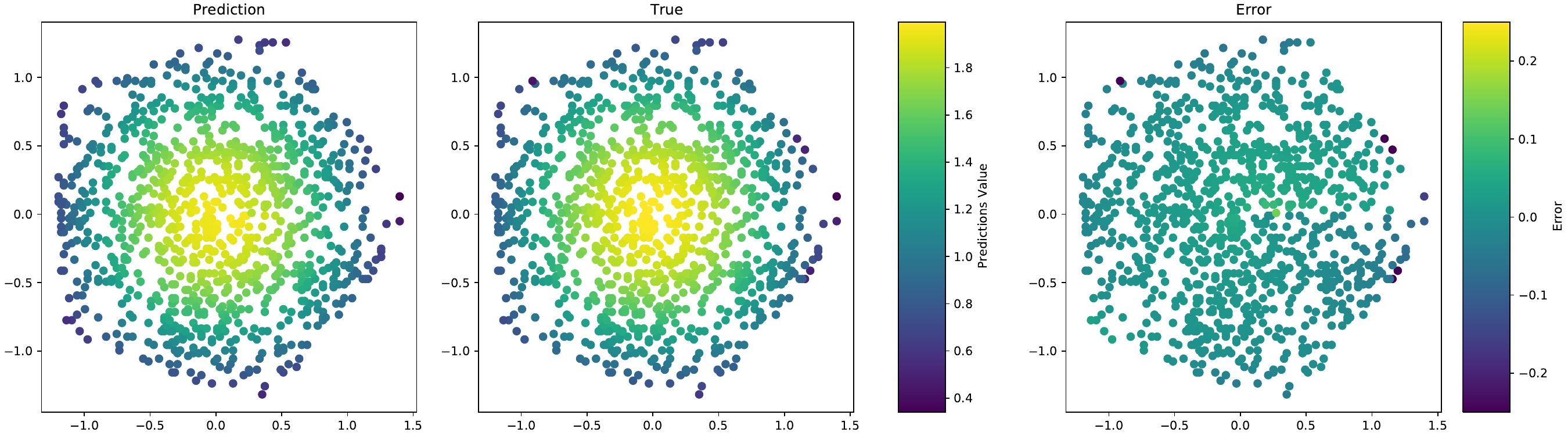} 
\caption{Results}
\label{fig:pplot2}
\end{figure}

We take the second and fourth boundary conditions (B2 and B4) as examples to show the result of our method, the predicted interior points, the true points, and the mean square error are shown in Fig. \ref{fig:plot2} and Fig. \ref{fig:pplot2}.

Table. \ref{tab:tab2} provides a comparative analysis with other methods across varying sample sizes (50, 100, and 200). We can find that our method FIE-NO in all boundary conditions give the best result, and the next is NIE and ANIE.

\subsubsection{2D Helmholtz equation}

\textbf{Dirichlet Boundary} For the 2D Helmholtz equation, the same set of Dirichlet Boundary conditions as the 2D Laplace equation is used, and we get the comparison result as the following Table. \ref{tab:tab3}:

\begin{table*}[h]
\centering
\caption{Results with Dirichlet Boundary of 2D Helmholtz equation(values in units of \(10^{-3}\))}
\begin{adjustbox}{width=\textwidth,center} 
\begin{tabular}{|c|c|c|c|c|c|c|c|c|c|c|c|c|c|}
\hline
\multirow{2}{*}{} & \multicolumn{3}{c|}{B1} & \multicolumn{3}{c|}{B2} & \multicolumn{3}{c|}{B3} & \multicolumn{3}{c|}{B4} \\ \cline{2-13}
                  & 50 & 100 & 200 & 50 & 100 & 200 & 50 & 100 & 200 & 50 & 100 & 200 \\ \hline
FIE-NO & \textbf{\textit{0.085}} & \textbf{\textit{0.057}} & \textbf{\textit{0.036}} & \textbf{\textit{0.080}} & \textbf{\textit{0.055}} & \textbf{\textit{0.033}} & \textbf{\textit{0.095}} & \textbf{\textit{0.064}} & \textbf{\textit{0.043}} & \textbf{\textit{0.146}} & \textbf{\textit{0.091}} & 
\textbf{\textit{0.062}}\\ \hline
NIE & 0.092 & 0.066 & 0.044 & 0.085 & 0.063 & 0.041 & 0.101 & 0.079 & 0.051 & 0.182 & 0.120 & 0.078 \\ \hline
ANIE & 0.088 & 0.061 & 0.039 & 0.082 & 0.056 & 0.034 & 0.099 & 0.071 & 0.049 & 0.166 & 0.109 & 0.075 \\ \hline
GO & 0.098 & 0.077 & 0.056 & 0.097 & 0.076 & 0.056 & 0.119 & 0.090 & 0.067 & 0.175 & 0.124 & 0.088 \\ \hline
GL & 0.099 & 0.080 & 0.059 & 0.103 & 0.082 & 0.061 & 0.114 & 0.083 & 0.062 & 0.180 & 0.129 & 0.097 \\ \hline
\end{tabular}
\end{adjustbox}
\label{tab:tab3}
\end{table*}

\noindent \textbf{Neumann Boundary} In this scenario, for the 2D Helmholtz equation, the same set of Neumann Boundary conditions as the 2D Laplace equation are used, we get the comparison result as the following Table. \ref{tab:tab4}. It can be seen the proposed FIE-NO consistently outperform other methods.

\begin{table*}[h]
\centering
\caption{Results with Neumann Boundary of 2D Helmholtz equation(values in units of \(10^{-3}\))}
\begin{adjustbox}{width=\textwidth,center} 
\begin{tabular}{|c|c|c|c|c|c|c|c|c|c|c|c|c|c|}
\hline
\multirow{2}{*}{} & \multicolumn{3}{c|}{B1} & \multicolumn{3}{c|}{B2} & \multicolumn{3}{c|}{B3} & \multicolumn{3}{c|}{B4} \\ \cline{2-13}
                  & 50 & 100 & 200 & 50 & 100 & 200 & 50 & 100 & 200 & 50 & 100 & 200 \\ \hline
FIE-NO & \textbf{\textit{0.085}} & \textbf{\textit{0.057}} & \textbf{\textit{0.036}} & \textbf{\textit{0.080}} & \textbf{\textit{0.055}} & \textbf{\textit{0.033}} & \textbf{\textit{0.095}} & \textbf{\textit{0.064}} & \textbf{\textit{0.043}} & \textbf{\textit{0.146}} & \textbf{\textit{0.091}} & 
\textbf{\textit{0.062}}\\ \hline
NIE & 0.092 & 0.066 & 0.044 & 0.085 & 0.063 & 0.041 & 0.101 & 0.079 & 0.051 & 0.182 & 0.120 & 0.078 \\ \hline
ANIE & 0.088 & 0.061 & 0.039 & 0.082 & 0.056 & 0.034 & 0.099 & 0.071 & 0.049 & 0.166 & 0.109 & 0.075 \\ \hline
GO & 0.098 & 0.077 & 0.056 & 0.097 & 0.076 & 0.056 & 0.119 & 0.090 & 0.067 & 0.175 & 0.124 & 0.088 \\ \hline
GL & 0.099 & 0.080 & 0.059 & 0.103 & 0.082 & 0.061 & 0.114 & 0.083 & 0.062 & 0.180 & 0.129 & 0.097 \\ \hline
\end{tabular}
\end{adjustbox}
\label{tab:tab4}
\end{table*}

\subsubsection{2D Darcy flow equation}

We set $ a(x, y) = 1 + 0.5 \sin\left(\sqrt{x^2 + y^2}\right) $ and $
f(x, y) = \sin\left(\sqrt{x^2 + y^2}\right)
$.

\textbf{Dirichlet Boundary} For the 2D Darcy flow equation, the same set of Dirichlet Boundary conditions as the 2D Laplace equation is used, and we get the comparison result as the following Table. \ref{tab:tab5}.

\begin{table*}[h]
\centering
\caption{Results with Dirichlet Boundary of Darcy flow equation(values in units of \(10^{-3}\))}
\begin{adjustbox}{width=\textwidth,center}
\begin{tabular}{|c|c|c|c|c|c|c|c|c|c|c|c|c|c|}
\hline
\multirow{2}{*}{} & \multicolumn{3}{c|}{B1} & \multicolumn{3}{c|}{B2} & \multicolumn{3}{c|}{B3} & \multicolumn{3}{c|}{B4} \\ \cline{2-13}
                  & 50 & 100 & 200 & 50 & 100 & 200 & 50 & 100 & 200 & 50 & 100 & 200 \\ \hline
FIE-NO & \textbf{\textit{0.085}} & \textbf{\textit{0.057}} & \textbf{\textit{0.036}} & \textbf{\textit{0.080}} & \textbf{\textit{0.055}} & \textbf{\textit{0.033}} & \textbf{\textit{0.095}} & \textbf{\textit{0.064}} & \textbf{\textit{0.043}} & \textbf{\textit{0.146}} & \textbf{\textit{0.091}} & 
\textbf{\textit{0.062}}\\ \hline
NIE & 0.092 & 0.066 & 0.044 & 0.085 & 0.063 & 0.041 & 0.101 & 0.079 & 0.051 & 0.182 & 0.120 & 0.078 \\ \hline
ANIE & 0.088 & 0.061 & 0.039 & 0.082 & 0.056 & 0.034 & 0.099 & 0.071 & 0.049 & 0.166 & 0.109 & 0.075 \\ \hline
GO & 0.098 & 0.077 & 0.056 & 0.097 & 0.076 & 0.056 & 0.119 & 0.090 & 0.067 & 0.175 & 0.124 & 0.088 \\ \hline
GL & 0.099 & 0.080 & 0.059 & 0.103 & 0.082 & 0.061 & 0.114 & 0.083 & 0.062 & 0.180 & 0.129 & 0.097 \\ \hline
\end{tabular}
\end{adjustbox}
\label{tab:tab5}
\end{table*}

\textbf{Neumann Boundary} In this scenario, the same set of Neumann Boundary conditions as the 2D Laplace equation are used, and we get the comparison result as the following Table. \ref{tab:tab6}.

Again, it can be seen from Table. \ref{tab:tab5} and Table. \ref{tab:tab6} that the proposed FIE-NO outperform the state-of-the-art methods consistently. 

\begin{table*}[h]
\centering
\caption{Results with Neumann Boundary of Darcy flow equation(values in units of \(10^{-3}\))}

\begin{adjustbox}{width=\textwidth,center}
\begin{tabular}{|c|c|c|c|c|c|c|c|c|c|c|c|c|c|}
\hline
\multirow{2}{*}{} & \multicolumn{3}{c|}{B1} & \multicolumn{3}{c|}{B2} & \multicolumn{3}{c|}{B3} & \multicolumn{3}{c|}{B4} \\ \cline{2-13}
                  & 50 & 100 & 200 & 50 & 100 & 200 & 50 & 100 & 200 & 50 & 100 & 200 \\ \hline
FIE-NO & \textbf{\textit{0.085}} & \textbf{\textit{0.057}} & \textbf{\textit{0.036}} & \textbf{\textit{0.080}} & \textbf{\textit{0.055}} & \textbf{\textit{0.033}} & \textbf{\textit{0.095}} & \textbf{\textit{0.064}} & \textbf{\textit{0.043}} & \textbf{\textit{0.146}} & \textbf{\textit{0.091}} & 
\textbf{\textit{0.062}}\\ \hline
NIE & 0.092 & 0.066 & 0.044 & 0.085 & 0.063 & 0.041 & 0.101 & 0.079 & 0.051 & 0.182 & 0.120 & 0.078 \\ \hline
ANIE & 0.088 & 0.061 & 0.039 & 0.082 & 0.056 & 0.034 & 0.099 & 0.071 & 0.049 & 0.166 & 0.109 & 0.075 \\ \hline
GO & 0.098 & 0.077 & 0.056 & 0.097 & 0.076 & 0.056 & 0.119 & 0.090 & 0.067 & 0.175 & 0.124 & 0.088 \\ \hline
GL & 0.099 & 0.080 & 0.059 & 0.103 & 0.082 & 0.061 & 0.114 & 0.083 & 0.062 & 0.180 & 0.129 & 0.097 \\ \hline
\end{tabular}
\end{adjustbox}
\label{tab:tab6}
\end{table*}

\subsubsection{Result analysis}

The study provides an in-depth analysis of the performance of the proposed neural operator approach compared to other methods, using the 2D Laplace, Helmholtz, and Darcy flow equation as test cases. It is shown that the error decreases as the sample size increases and highlights the robustness of our proposed method. The better performance proves the proposed model's effectiveness, making it a promising tool for complex boundary value problems in computational physics and engineering.

\section{Related work}

In this section, we provide a discussion of existing literature related to solving Data-driven Boundary Value Problems, with a particular focus on the challenges posed by irregular boundaries. 

The integration of machine learning into this domain has been a significant step towards addressing these challenges. Such as, Fang. et al. use neural operators to learn boundary value problems of specific functions \cite{learningBoundary}, this work needs Green's functions of specific functions. Wang. et al. design a neural operator with transformers to solve BIEs for Elliptic PDEs \cite{beno}. This synergy offers a reduction in computational demands and enhances the ability to handle complex geometries and unbounded domains with improved accuracy. Besides, the classic Green's functions provide a foundational bridge between traditional analytical techniques and modern computational strategies due to its integral kernels, making them indispensable in many applications. Green's functions are indeed a crucial tool in solving boundary value problems, often used to express solutions to differential equations given certain boundary conditions \cite{greenbook}. Therefore, data-driven learning of Green’s function \cite{greenlearning} offers a powerful tool for addressing computational challenges in BIEs, which proposes a deep learning approach to model both the Green’s function and the homogeneous solution.

In the evolving landscape of computational methods for solving Boundary Value Problems (BVPs), Physics-Informed Neural Networks (PINNs) represent a significant innovation \cite{PIdon}. PINNs incorporate boundary conditions directly into the training process by penalizing the loss function based on deviations from these constraints. This method ensures that the solutions adhere to both the governing differential equations and the boundary conditions throughout the training phase. In contrast, the FIE-NO method presented in this work integrates boundary conditions through the mathematical formulation of Fredholm Integral Equations. Unlike PINNs, which adjust the loss function to enforce boundary adherence, FIE-NO embeds this information within the integral equations themselves, offering a seamless integration that can be inherently more adaptable to complex geometries and diverse boundary conditions. This fundamental difference underlines the potential of FIE-NO to generalize across varied scenarios without the need for specific loss functions, thus simplifying the implementation while enhancing the solution's robustness to changes in boundary specifications.

\section{Conclusion}

This paper proposed a new method to solve data-driven boundary value problems. By integrating RFF and FIE into deep learning, we create a physics-guided machine learning method.
The proposed learning method was validated with two numerical examples and compared with the state-of-the-art algorithms. It was shown that the proposed method outperformed existing methods on data-driven boundary value problems.

\bibliographystyle{alpha}
\bibliography{sample}

\newcommand{\etalchar}[1]{$^{#1}$}
\begin{thebibliography}{LKA{\etalchar{+}}20b}

\bibitem[ACLP24]{aldirany2024operator}
Ziad Aldirany, R{\'e}gis Cottereau, Marc Laforest, and Serge Prudhomme.
\newblock Operator approximation of the wave equation based on deep learning of green's function.
\newblock {\em Computers \& Mathematics with Applications}, 159:21--30, 2024.

\bibitem[BET22]{greenlearning}
Nicolas Boull{\'e}, Christopher~J Earls, and Alex Townsend.
\newblock Data-driven discovery of green’s functions with human-understandable deep learning.
\newblock {\em Scientific reports}, 12(1):4824, 2022.

\bibitem[BPBP23]{PIdon}
R{\"u}diger Brecht, Dmytro~R Popovych, Alex Bihlo, and Roman~O Popovych.
\newblock Improving physics-informed deeponets with hard constraints.
\newblock {\em arXiv preprint arXiv:2309.07899}, 2023.

\bibitem[BT23]{boulleoperator}
Nicolas Boull{\'e} and Alex Townsend.
\newblock A mathematical guide to operator learning.
\newblock {\em arXiv preprint arXiv:2312.14688}, 2023.

\bibitem[FWP23]{learningBoundary}
Zhiwei Fang, Sifan Wang, and Paris Perdikaris.
\newblock Learning only on boundaries: a physics-informed neural operator for solving parametric partial differential equations in complex geometries.
\newblock {\em arXiv preprint arXiv:2308.12939}, 2023.

\bibitem[HW08]{BIEbook}
George~C Hsiao and Wolfgang~L Wendland.
\newblock {\em Boundary integral equations}.
\newblock Springer, 2008.

\bibitem[KLL{\etalchar{+}}23]{NO}
Nikola Kovachki, Zongyi Li, Burigede Liu, Kamyar Azizzadenesheli, Kaushik Bhattacharya, Andrew Stuart, and Anima Anandkumar.
\newblock Neural operator: Learning maps between function spaces with applications to pdes.
\newblock {\em Journal of Machine Learning Research}, 24(89):1--97, 2023.

\bibitem[Lad13]{bvp_book}
Olga~Aleksandrovna Ladyzhenskaya.
\newblock {\em The boundary value problems of mathematical physics}, volume~49.
\newblock Springer Science \& Business Media, 2013.

\bibitem[Lan51]{FIE}
Louis Landweber.
\newblock An iteration formula for fredholm integral equations of the first kind.
\newblock {\em American journal of mathematics}, 73(3):615--624, 1951.

\bibitem[LJP{\etalchar{+}}21]{deeponet}
Lu~Lu, Pengzhan Jin, Guofei Pang, Zhongqiang Zhang, and George~Em Karniadakis.
\newblock Learning nonlinear operators via deeponet based on the universal approximation theorem of operators.
\newblock {\em Nature machine intelligence}, 3(3):218--229, 2021.

\bibitem[LKA{\etalchar{+}}20a]{FNO}
Zongyi Li, Nikola Kovachki, Kamyar Azizzadenesheli, Burigede Liu, Kaushik Bhattacharya, Andrew Stuart, and Anima Anandkumar.
\newblock Fourier neural operator for parametric partial differential equations.
\newblock {\em arXiv preprint arXiv:2010.08895}, 2020.

\bibitem[LKA{\etalchar{+}}20b]{gkn}
Zongyi Li, Nikola Kovachki, Kamyar Azizzadenesheli, Burigede Liu, Kaushik Bhattacharya, Andrew Stuart, and Anima Anandkumar.
\newblock Neural operator: Graph kernel network for partial differential equations.
\newblock {\em arXiv preprint arXiv:2003.03485}, 2020.

\bibitem[LOO22]{bvplearning}
Winfried L{\"o}tzsch, Simon Ohler, and Johannes~S Otterbach.
\newblock Learning the solution operator of boundary value problems using graph neural networks.
\newblock {\em arXiv preprint arXiv:2206.14092}, 2022.

\bibitem[LZZ{\etalchar{+}}24]{lu2024fast}
Zibo Lu, Yuanye Zhou, Yanbo Zhang, Xiaoguang Hu, Qiao Zhao, and Xuyang Hu.
\newblock A fast general thermal simulation model based on multi-branch physics-informed deep operator neural network.
\newblock {\em Physics of Fluids}, 36(3), 2024.

\bibitem[RPK19]{pinn}
Maziar Raissi, Paris Perdikaris, and George~E Karniadakis.
\newblock Physics-informed neural networks: A deep learning framework for solving forward and inverse problems involving nonlinear partial differential equations.
\newblock {\em Journal of Computational physics}, 378:686--707, 2019.

\bibitem[SH11]{greenbook}
Ivar Stakgold and Michael~J Holst.
\newblock {\em Green's functions and boundary value problems}.
\newblock John Wiley \& Sons, 2011.

\bibitem[SLG21]{salvi2021neural}
Cristopher Salvi, Maud Lemercier, and Andris Gerasimovics.
\newblock Neural stochastic partial differential equations.
\newblock {\em arXiv preprint arXiv}, 2110:106, 2021.

\bibitem[WLD{\etalchar{+}}24]{beno}
Haixin Wang, Jiaxin Li, Anubhav Dwivedi, Kentaro Hara, and Tailin Wu.
\newblock Beno: Boundary-embedded neural operators for elliptic pdes.
\newblock {\em arXiv preprint arXiv:2401.09323}, 2024.

\bibitem[ZFCvD22]{2022NIE}
Emanuele Zappala, Antonio Henrique de~Oliveira Fonseca, Josue~Ortega Caro, and David van Dijk.
\newblock Neural integral equations.
\newblock {\em arXiv preprint arXiv:2209.15190}, 2022.

\end{thebibliography}

\newpage
\section*{Appendix A: Detailed Derivation from Equation 9 to Equation 12}

The left hand-side of Equation 9 is rewritten below:
\[
\int_S \cos(\omega_i^\top t + b_i) \phi(t) \, dS(t)
\]

Expanding \(\phi(t)\) as a Fourier series:
\[
\phi(t) = \sum_n \left( a_n \cos(nt) + b_n \sin(nt) \right)
\]

Substituting \(\phi(t)\), we get the full Equation 9:
\[\int_S \cos(\omega_i^\top t + b_i) \phi(t) \, dS(t) = 
\int_S \cos(\omega_i^\top t + b_i) \left( \sum_n a_n \cos(nt) + b_n \sin(nt) \right) \, dS(t)
\]

Using the cosine angle addition formula:
\[
\cos(\omega_i^\top t + b_i) = \cos(\omega_i^\top t) \cos(b_i) - \sin(\omega_i^\top t) \sin(b_i)
\]

Substituting this into the right hand-side of Equation 9:
\[
\int_S \left( \cos(\omega_i^\top t) \cos(b_i) - \sin(\omega_i^\top t) \sin(b_i) \right) \left( \sum_n a_n \cos(nt) + b_n \sin(nt) \right) \, dS(t)
\]

Expanding the above integral into four parts:
\[
\cos(b_i) \sum_n a_n \int_S \cos(\omega_i^\top t) \cos(nt) \, dS(t) - \cos(b_i) \sum_n b_n \int_S \cos(\omega_i^\top t) \sin(nt) \, dS(t)
\]
\[
- \sin(b_i) \sum_n a_n \int_S \sin(\omega_i^\top t) \cos(nt) \, dS(t) + \sin(b_i) \sum_n b_n \int_S \sin(\omega_i^\top t) \sin(nt) \, dS(t)
\]

Utilizing orthogonality to simplify integrals:
\[
\begin{aligned}
& \int_S \cos(\omega_i^\top t) \cos(nt) \, dS(t) \text{ is zero when } n \neq \omega_i . \\
& \int_S \cos(\omega_i^\top t) \sin(nt) \, dS(t) \text{ is always zero because } \cos \text{ and } \sin \text{ are orthogonal.} \\
& \int_S \sin(\omega_i^\top t) \cos(nt) \, dS(t) \text{ is always zero because } \sin \text{ and } \cos \text{ are orthogonal.} \\
& \int_S \sin(\omega_i^\top t) \sin(nt) \, dS(t) \text{ is zero when }  n \neq \omega_i .
\end{aligned}
\]

It can be seen the only the non-zero terms, when \(n = \omega_i \), remain:
\[
\cos(b_i) a_{\omega_i} \int_S \cos^2(\omega_i^\top t) \, dS(t) - \sin(b_i) b_{\omega_i} \int_S \sin^2(\omega_i^\top t) \, dS(t)
\]

Using the trigonometric identity for squares:
\[
\cos^2(x) = \frac{1 + \cos(2x)}{2}
\]
\[
\sin^2(x) = \frac{1 - \cos(2x)}{2}
\]

Substituting these into the integrals:
\[
\cos(b_i) a_{\omega_i} \int_S \frac{1 + \cos(2\omega_i^\top t)}{2} \, dS(t) - \sin(b_i) b_{\omega_i} \int_S \frac{1 - \cos(2\omega_i^\top t)}{2} \, dS(t)
\]

Separating the integral terms, we get:
\[
\cos(b_i) a_{\omega_i} \left( \frac{1}{2} \int_S dS(t) + \frac{1}{2} \int_S \cos(2\omega_i^\top t) \, dS(t) \right) - \sin(b_i) b_{\omega_i} \left( \frac{1}{2} \int_S dS(t) - \frac{1}{2} \int_S \cos(2\omega_i^\top t) \, dS(t) \right)
\]

Since \(\cos(2\omega_i^\top t)\) integrates to zero over the entire boundary, which represents \([0, 2\pi]\):
\[
\int_S \cos(2\omega_i^\top t) \, dS(t) = 0
\]

Thus, the above expression simplifies to:
\[
\cos(b_i) a_{\omega_i} \frac{1}{2} \int_S dS(t) - \sin(b_i) b_{\omega_i} \frac{1}{2} \int_S dS(t)
\]

Extracting the common factor and defining a constant \(c\):
\[
c = \frac{1}{2} \left( \cos(b_i) a_{\omega_i} - \sin(b_i) b_{\omega_i} \right)
\]

Therefore, the final expression simplies to:
\[
\int_S \cos(\omega_i^\top t + b_i) \phi(t) \, dS(t) = c \int_S dS(t)
\]

\end{document}